\documentclass{article}

 \usepackage{todonotes}
 \usepackage[shortlabels]{enumitem}
 \usepackage{amsmath}
 \usepackage[preprint]{neurips_2025}

\usepackage[utf8]{inputenc} % allow utf-8 input
\usepackage[T1]{fontenc}    % use 8-bit T1 fonts
\usepackage{hyperref}       % hyperlinks
\usepackage{url}            % simple URL typesetting
\usepackage{booktabs}       % professional-quality tables
\usepackage{amsfonts}       % blackboard math symbols
\usepackage{nicefrac}       % compact symbols for 1/2, etc.
\usepackage{microtype}      % microtypography
\usepackage{xcolor}         % colors

\title{Causality $\neq$ Decodability, and Vice Versa: Lessons from Interpreting Counting ViTs}

\author{%
  Lianghuan Huang \\
  University of Pennsylvania\\
  Philadelphia, PA 19104 \\
  \texttt{leoh@sas.upenn.edu} \\
  \And
  Yingshan Chang \\
  Carnegie Mellon University \\
  Pittsburgh, PA 15213 \\
  \texttt{yingshac@andrew.cmu.edu} \\
}

\begin{document}

\maketitle

\begin{abstract}
Mechanistic interpretability seeks to uncover how internal components of neural networks give rise to predictions. A persistent challenge, however, is disentangling two often conflated notions: decodability—the recoverability of information from hidden states—and causality—the extent to which those states functionally influence outputs. In this work, we investigate their relationship in vision transformers (ViTs) fine-tuned for object counting. Using activation patching, we test the causal role of spatial and CLS tokens by transplanting activations across clean–corrupted image pairs. In parallel, we train linear probes to assess the decodability of count information at different depths. Our results reveal systematic mismatches: middle-layer object tokens exert strong causal influence despite being weakly decodable, whereas final-layer object tokens support accurate decoding yet are functionally inert. Similarly, the CLS token becomes decodable in mid-layers but only acquires causal power in the final layers. These findings highlight that decodability and causality reflect complementary dimensions of representation—what information is present versus what is used—and that their divergence can expose hidden computational circuits.
\end{abstract}

\section{Introduction}

Mechanistic interpretability seeks to uncover how internal components of neural networks contribute to predictions, moving beyond aggregate performance metrics toward causal understanding of model behavior~\cite{bereska2024mechanisticinterpretabilityaisafety, olahMech, feng2024languagemodelsbindentities, hanna2023doesgpt2computegreaterthan, nanda2023progressmeasuresgrokkingmechanistic,wu2024interpretabilityscaleidentifyingcausal, Joseph_Nanda_2024, lepori2024doorsperceptionvisiontransformers, liu2025mechanisticinterpretabilitymeets}. A central challenge is distinguishing between two notions often conflated in practice: the decodability of information from representations, and the causal use of that information by the model. Decodability methods, such as linear probing, test whether a variable of interest can be recovered from hidden states~\cite{belinkov2021probingclassifierspromisesshortcomings}. Causal methods, such as activation patching, instead test whether modifying activations changes the model’s outputs~\cite{heimersheim2024useinterpretactivationpatching, zhang2024bestpracticesactivationpatching}. While both approaches provide valuable perspectives, it remains unclear how they align—or diverge—across layers and token types in large models.

This distinction is especially pertinent for vision transformers (ViTs), whose predictions arise from the interaction of local patch embeddings and a global classification (CLS) token. In ViTs, local patches may contain object-specific information, while the CLS token aggregates global scene evidence. Yet whether information contained in these representations is actually used in making predictions is less well understood. For example, a token may carry highly decodable features but exert no influence on the output, or conversely, a token with weakly decodable information may nonetheless causally drive predictions when perturbed. Understanding this gap is crucial for accurately characterizing what ViTs represent and how they compute.

In this work, we investigate the relationship between decodability and causality in a vision transformer fine-tuned for object counting~\cite{chang2024languagemodelsneedinductive, kajic2022probing, kajić2025evaluatingnumericalreasoningtexttoimage}. Using activation patching, we transplant hidden activations from clean and corrupted image pairs to test which tokens influence predictions at different depths. In parallel, we train linear probes on object patches, CLS tokens, and background patches to assess their decodability. Comparing the two perspectives reveals systematic mismatches. Middle-layer object tokens, though weakly decodable, exert strong causal influence when patched. By contrast, final-layer object tokens support highly accurate decoding yet are functionally inert, with predictions unaffected by patching. Similarly, CLS tokens become decodable in the middle layers but only acquire causal power in the final layers.

Our findings highlight that decodability and causality are not interchangeable lenses on model behavior. Instead, they reflect complementary dimensions of representation: what information is present, and what information is used. By demonstrating their divergence in a concrete setting, we argue that both are necessary for a comprehensive interpretability analysis, and that mismatches between them may reveal hidden computational circuits.

\section{Activation Patching for Causality}
\label{activation_patching}

Activation patching~\cite{heimersheim2024useinterpretactivationpatching, zhang2024bestpracticesactivationpatching} is a causal interpretability method that tests whether specific activations are used by the model to make predictions. Given a source input $x_s$ and a target input $x_t$, we forward each through the model to obtain hidden states $h^l_s, h^l_t$ at layer $l$. We then form a patched run by replacing part of the target activations with those from the source:

$$
\tilde{h}^l_t = h^l_t \;\;\text{with}\;\; h^l_{t,i} \leftarrow h^l_{s,i},
$$

where $i$ indexes the token or component being patched. Continuing the forward pass from $\tilde{h}^l_t$ yields a new output $\tilde{f}(x_t)$. If $\tilde{f}(x_t)$ shifts toward $f(x_s)$, then the patched activations causally influence the prediction.

\begin{figure}[h]
    \centering
    \includegraphics[width=\linewidth]{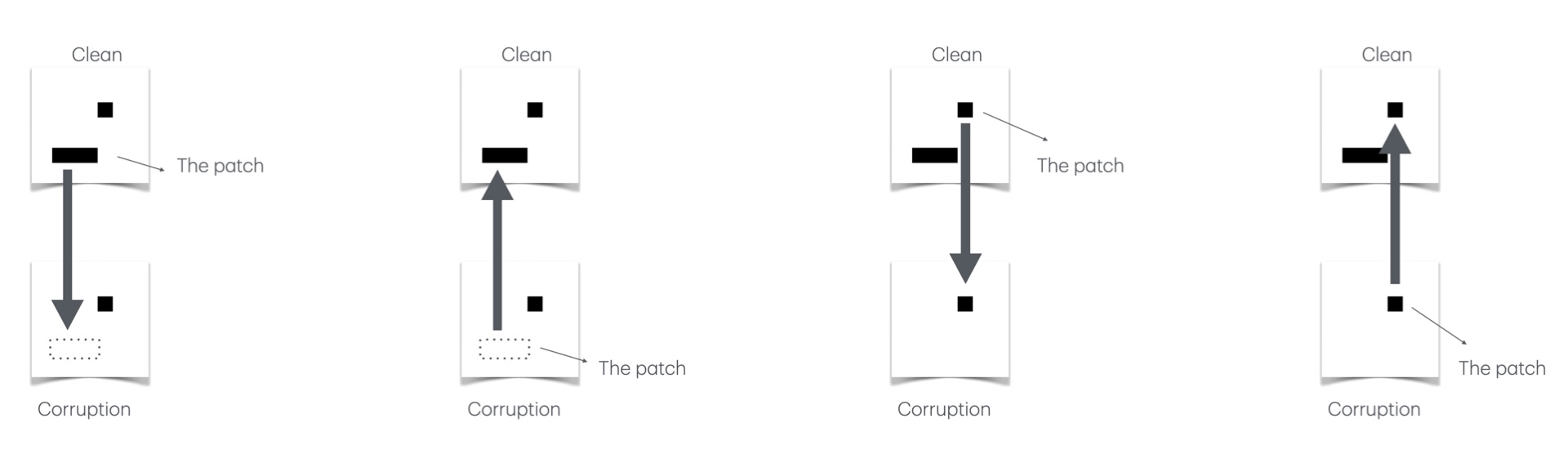}
    \caption{Activation patching experiments, corresponding to experiments 1 through 4, where object A is the $1\times 1$ square and object B is the $1\times 3$ rectangle.}
    \label{fig:activation_patching}
\end{figure}

As an illustrative example of our activation patching setup, we consider an image containing two objects, A and B (the clean image). We construct a corrupted counterpart by removing object B. When passed through the counting-finetuned vision transformer, the clean image yields a prediction of 2, while the corrupted image yields a prediction of 1, as expected (see Appendix~\ref{section:exp-details} for details on the model and dataset). We then perform the following activation patching experiments (Figure~\ref{fig:activation_patching}):
\begin{enumerate}
    \item Patch the activation of object B from the clean run into the corrupted run.  
    \item Patch the activation of the empty patches in the corrupted run (corresponding to the patch locations of object B in the clean run) into the clean run.  
    \item Patch the activation of object A from the clean run into the corrupted run.  
    \item Patch the activation of object A from the corrupted run into the clean run.  
    \item Patch the CLS token from the clean run into the corrupted run.  
    \item Patch the CLS token from the corrupted run into the clean run.  
\end{enumerate}

\begin{figure}[h]
    \centering
    \includegraphics[width=0.8\linewidth]{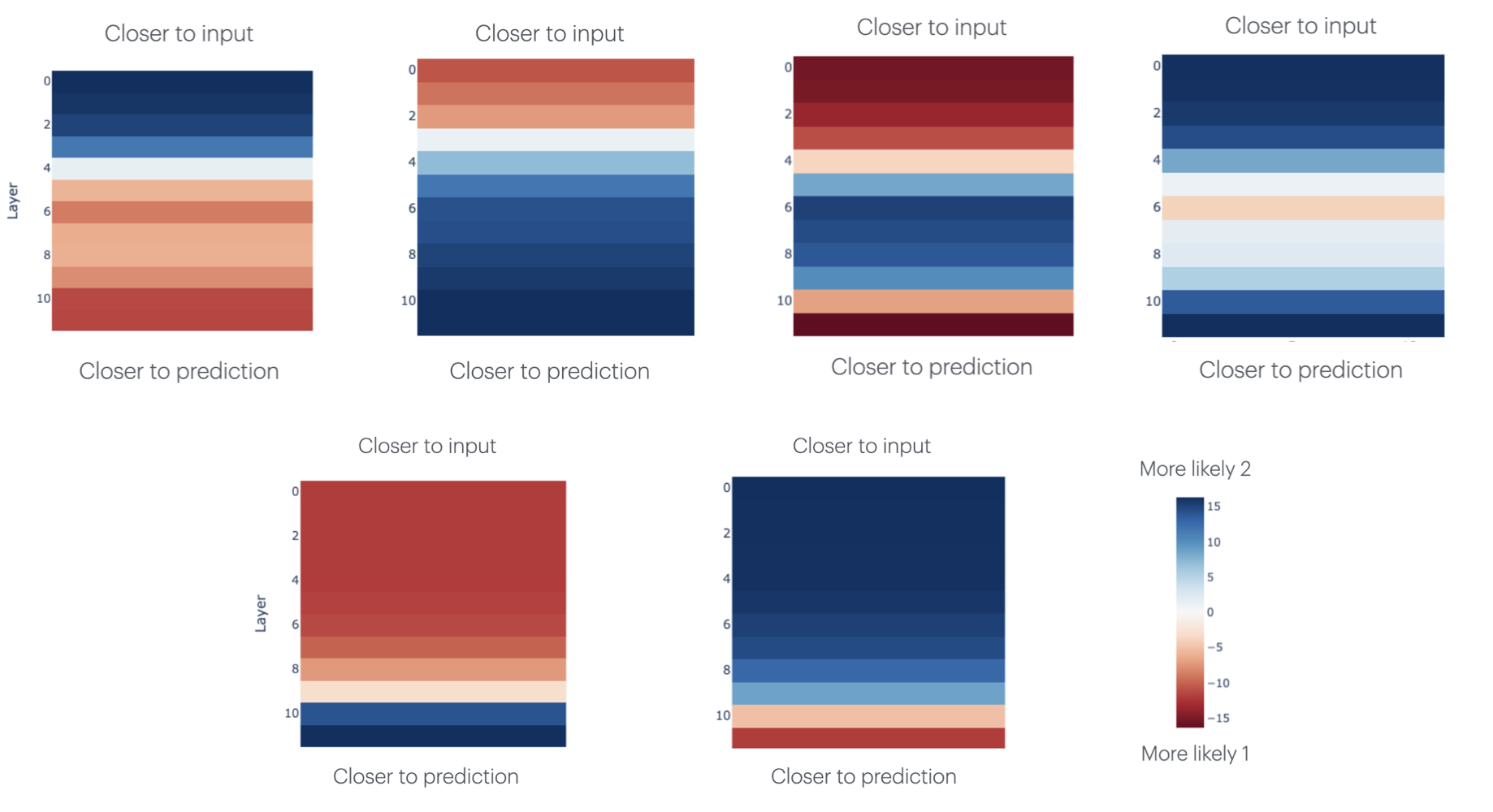}
    \caption{Activation patching results, corresponding to experiments 1 through 6 in left-right, top-bottom order.}
    \label{fig:ap-results}
\end{figure}

In our setup, activations are patched layer-wise: an activation from a given layer in the source run is transplanted into the corresponding layer of the target run. To assess how information at different depths influences model behavior, we measure the effect of patching by computing the logit difference between the model’s predictions for class 2 versus class 1 as a function of the patched layer. The results are shown in Figure~\ref{fig:ap-results}.

In experiment 1, patching a token corresponding to an “additional object” flips the prediction from 1 to 2, but only when the patching is performed in early layers. Likewise, in experiment 2, patching an “empty” token that effectively occludes object B flips the prediction from 2 to 1, again restricted to early layers. These results suggest that early layers in a ViT process local patch information in a way that remains directly relevant to the model’s predictions.

One might expect that transplanting an activation corresponding to an object token into a location where another object is already present would leave the model’s prediction unchanged. Surprisingly, our experiments 3 and 4 reveal otherwise. In particular, when we patch a middle-layer activation of object A from a run with two objects into the corresponding token of a run with one object, the prediction shifts from 1 (the count in the target image) to 2 (the count in the source image). Conversely, patching object A’s activation in the middle layers from the one-object run into the two-object run causes the prediction to flip from 2 to 1, again aligning with the source. These findings indicate that object-containing spatial patches in the middle layers encode some information relevant to the global count of the image, and that this information can be transferred to the target run through patching, despite the ostensible count being unchanged.

In experiments 5 and 6, we patch the CLS token. As shown in Figure~\ref{fig:ap-results}, this intervention alters the prediction in the expected manner—flipping the target count to match the source count—only when applied in the final layers.

 Note that the effects above are observed consistently across multiple clean–corrupted image pairs with varying object counts (see Appendix~\ref{section:ap-more}).To further understand the information contained in the spatial and CLS patches, we next conduct a series of linear probing experiments on these tokens.

\section{Linear Probing for Decodability}
\begin{figure}[h]
    \centering
    \includegraphics[width=0.5\linewidth]{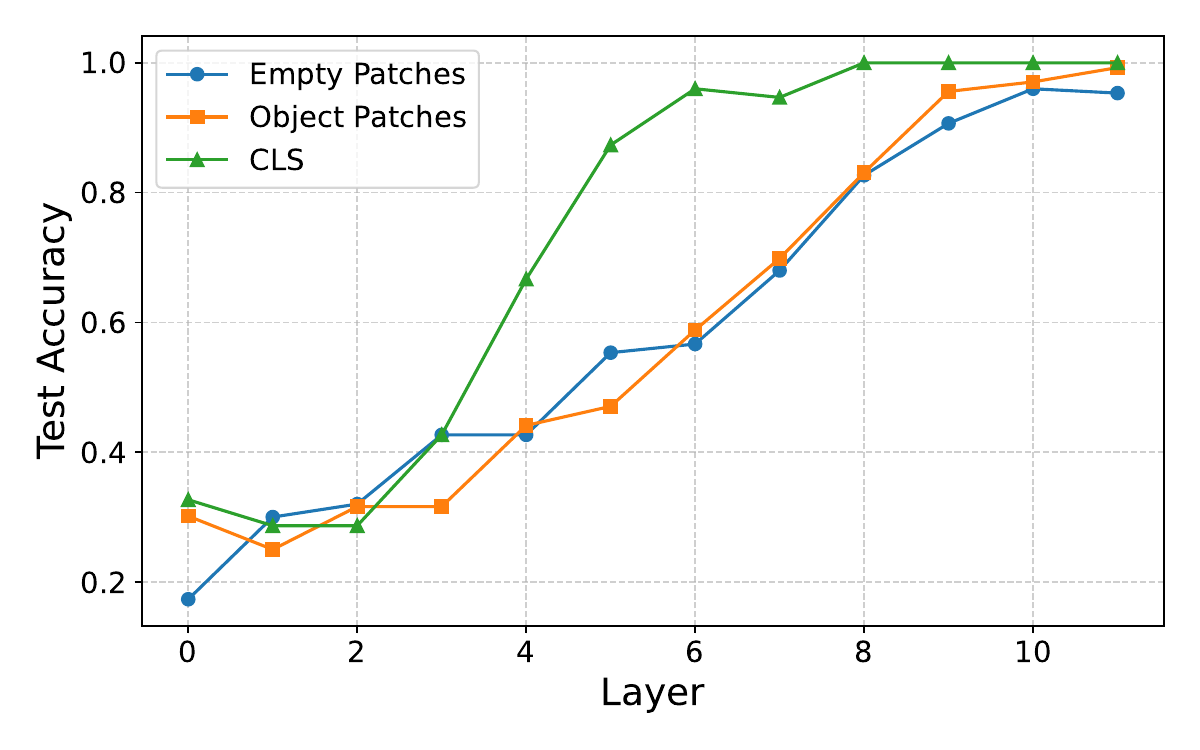}
    \caption{Test accuracy of linear probes.}
    \label{fig:lp}
\end{figure}
We perform linear probing on three categories of tokens: spatial patches containing objects, CLS tokens, and background patches (used as a baseline). For each image, we randomly sample one token from each category and train a linear classification probe with cross-entropy loss on the training split, reporting test accuracy on the held-out set. As shown in Figure~\ref{fig:lp}, probing accuracies in the early layers are uniformly low, indicating that the model is primarily engaged in local feature processing at this stage. Consistently, as demonstrated in experiments 1 and 2 above, patching local object tokens is still effective in altering predictions during these early layers.

In the middle layers, probe accuracies on spatial patches increase steadily, while the accuracy on the CLS token rises sharply to above 90\%. Importantly, in the same middle layers where prediction flipping was observed in experiments 3 and 4, the object patches themselves do not encode precise information about the global count, as indicated by the probes. It is only in the final layers that object patches achieve accuracies above 90\%; however, at this stage, patching object tokens no longer alters the prediction. These findings highlight our central argument: \textbf{decodability and causality are not equivalent}. Middle-layer object tokens exhibit causal influence on predictions despite being only weakly decodable, whereas final-layer object tokens support accurate decoding of count information yet lack causal influence on the model’s outputs.

Examining the CLS token, its probe accuracy reaches the 90\% level by the middle layers. Yet, as shown in experiments 5 and 6, patching the CLS token at this stage does not alter predictions—\textbf{again illustrating that decodability $\neq$ causality}. Only in the final layers does CLS token patching successfully flip the prediction.

The observed layer-wise mismatch between decodability and causality can suggest a plausible account of how the attention mechanism transports the count information, and eventually sends it to the CLS token: In the early layers, predictions are causally driven by local object tokens, consistent with the low probe accuracy of global information at this stage. By the middle layers, object tokens themselves contain only weak traces of the global count, yet attention may be actively \textit{reading} from them to shape the model’s output—explaining why patching these tokens can flip predictions. At the same time, the CLS token already encodes reasonably accurate count information, but attention may still be \textit{writing} new signals from spatial tokens into it, diminishing its causal role when patched. In the final layers, object and background tokens exhibit highly decodable count information, but attention appears no longer to propagate it into the CLS token. As a result, patching them has little effect, suggesting that the CLS token has already settled into its final prediction—consistent with its dominant causal influence at this stage. We emphasize, however, that this interpretation is preliminary. Our goal is to demonstrate that the mismatch between decodability and causality is systematic and can potentially expose hidden circuits—ones that implement computations more complex than standard associative memory or key–value retrieval~\cite{doi:10.1073/pnas.79.8.2554, olah2020zoom, olsson2022context, wang2022interpretabilitywildcircuitindirect}.

\section{Discussion}
\label{others}
Our results show that decodability and causality can diverge in systematic ways. A token may contain decodable information while functionally inert, or exert causal influence despite weak probe accuracy.

Probing alone can either overstate or understate functional roles: final-layer object tokens yield high decoding accuracy for counts, yet patching reveals they no longer affect predictions; conversely, middle-layer object tokens flip predictions when patched even though probes recover little count information. On the other hand, causal analysis alone can also mislead: patching identifies which tokens influence predictions, but not whether they do so by carrying reliable task information or by transmitting intermediate signals. Without probing, causal influence remains ambiguous.

Taken together, these findings show that decodability and causality vary dynamically across layers, and neither perspective alone is sufficient to capture how representations are used. A comprehensive interpretability analysis requires asking both: what information is present?
and what information is used?
\newpage
\appendix
\section{Activation Patching on Additional Pairs}
\label{section:ap-more}

\begin{figure}[h]
    \centering
    \includegraphics[width=0.5\linewidth]{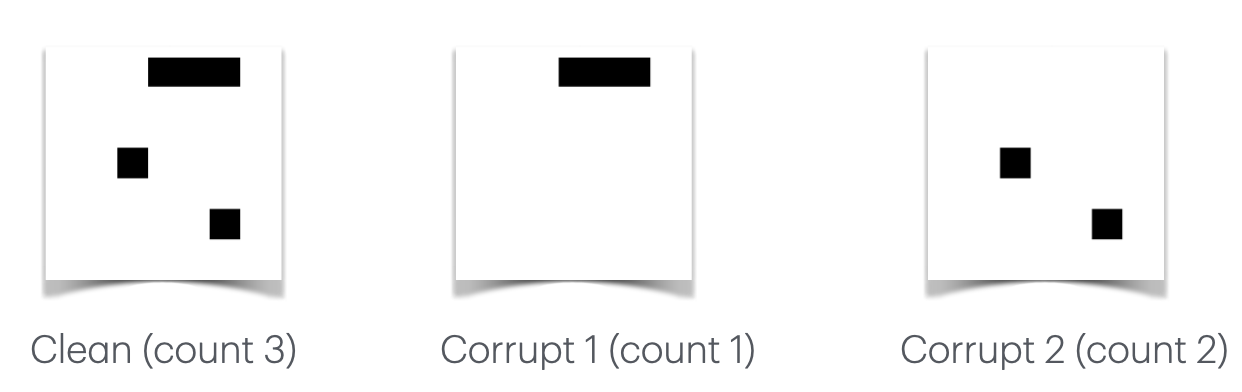}
    \caption{Activation patching with 3 objects}
    \label{fig:ap1}
\end{figure}

Here we provide further examples of activation patching with more objects. In the first case, the clean image contained three objects: one $1\times3$ rectangle and two $1\times1$ squares. Two corrupted versions were created. Corrupt 1 removed both squares, leaving only the rectangle (count = 1). Corrupt 2 removed the rectangle, leaving the two squares (count = 2).

For Corrupt 1, patching the rectangle token from the clean run restored the prediction from 1 to 2 when applied at layers 7–8 (0-indexed), but never recovered the full count of 3. Even when the rectangle was patched together with one of the square tokens, the prediction still only restored to 2.

For Corrupt 2, patching in both square tokens from the clean run restored the prediction from 2 to 3, when applied at layers 6–9 (0-indexed).

\begin{figure}[h]
\centering
\includegraphics[width=0.3\linewidth]{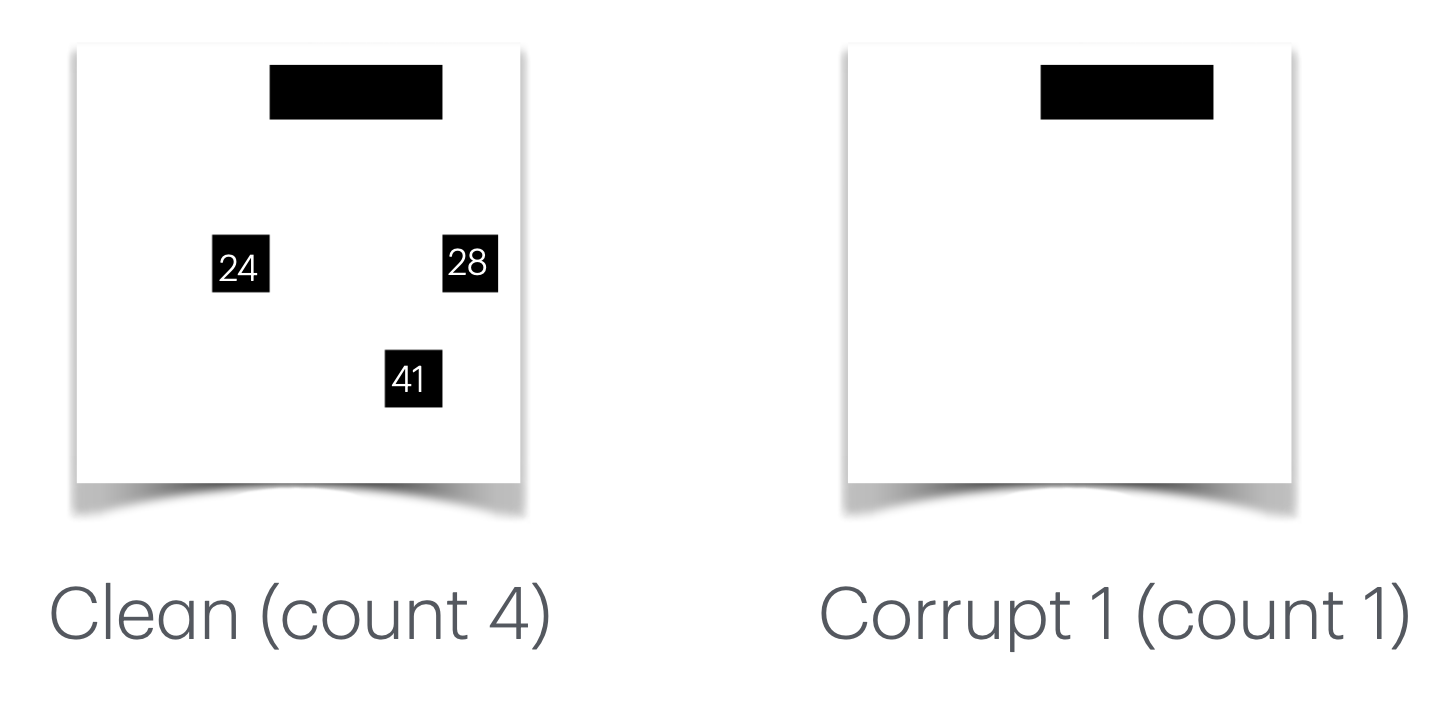}
\caption{Activation patching with 4 objects.}
\label{fig:ap2}
\end{figure}

Another example is a clean image with four objects (one $1\times3$ rectangle and three $1\times1$ squares) and a corrupted version containing only the rectangle (count = 1).

For this corrupted run, patching in the rectangle token from the clean run restored the prediction to 2 at layers 6–8 (0-indexed), but the effect diminished to 1 again in layers 9–11. Patching the rectangle together with one square temporarily restored the prediction to 3 at layers 6–7, fell to 2 at layer 8, and returned to 1 at layers 9–11. Finally, patching the rectangle with any two squares never restored the full count of 4.

These additional results reinforce the finding that \textbf{patching in the middle layers can influence predictions}. At the same time, they corroborate the probing conclusion that \textbf{count information in middle-layer object tokens is inaccurate}: predictions can be swayed in the direction of the source image, but they do not reliably recover the correct total count.

We conducted activation patching on a total of 20 pairs with different labels, and confirm that the patterns above is general.

\section{Experimental Details}
\label{section:exp-details}
\paragraph{Model}
 We use a standard Vision Transformer (ViT-B/32) architecture with 12 layers, 12 attention heads, and hidden dimension 768. The model is initialized from ImageNet-21k pretraining and ImageNet-1k fine-tuning~\cite{steiner2022trainvitdataaugmentation}, and we further fine-tune it on a synthetic object counting dataset. Fine-tuning is performed with a 10-way classification head (predicting counts from 1–10) using cross-entropy loss, Adam optimizer, learning rate $3\times 10^{-4}$, batch size 8192, and training for 250 epochs. Both training and testing accuracies reached 100\% after 225 epochs. All experiments are conducted with the same model checkpoint at the 250 epoch. 
\paragraph{Dataset}
Our synthetic dataset consists of images containing two object types: $1 \times 1$ squares and $1 \times 3$ rectangles. To facilitate activation patching, all objects are aligned with the patch grid of the ViT: each patch is either fully occupied (black) or left entirely empty. Object placement is randomized across images. For each target count from 1 to 10, we generate 100 images, resulting in a balanced distribution across counts. The dataset is split into training and test sets using a 75–25 ratio. 
\paragraph{Activation Patching}
We implement activation patching on vision transformers with adaptations of ViT-Prisma, an open-source library for vision transformer interpretability~\cite{ joseph2023vit, joseph2025prismaopensourcetoolkit}.

%%%%%%%%%%%%%%%%%%%%%%%%%%%%%%%%%%%%%%%%%%%%%%%%%%%%%%%%%%%%
\newpage

\bibliographystyle{plain} % We choose the "plain" reference style
\bibliography{refs} % Entries are in the refs.bib file

\end{document}